\documentclass{article}

     \usepackage[preprint,nonatbib]{neurips_2020}

\usepackage[utf8]{inputenc} 
\usepackage[T1]{fontenc}    
\usepackage{hyperref}       
\usepackage{url}            
\usepackage{booktabs}       
\usepackage{amsfonts}       
\usepackage{nicefrac}       
\usepackage{microtype}      
\usepackage{floatrow}
\usepackage{graphicx}
\usepackage{float}
\usepackage{subfigure}
\usepackage{amsmath}
\usepackage{algorithm}
\usepackage{algorithmic}
\usepackage{amssymb}
\usepackage{multirow}
\usepackage{bm}
\usepackage[]{caption2} 
\renewcommand{\thesubfigure}{(\roman{subfigure})}
\makeatletter \renewcommand{\@thesubfigure}{\thesubfigure \space}
\renewcommand{\p@subfigure}{} \makeatother

\usepackage{tikz}
\usepackage{etoolbox}
\newcommand{\circled}[2][]{\tikz[baseline=(char.base)]
	{\node[shape = circle, draw, inner sep = 1pt]
		(char) {\phantom{\ifblank{#1}{#2}{#1}}};%
		\node at (char.center) {\makebox[0pt][c]{#2}};}}
\robustify{\circled}

\title{Loosely Coupled Federated Learning Over Generative Models}

\author{%
  Shaoming Song \\
  Huawei Noah’s Ark Lab\\
  \texttt{shaoming.song@huawei.com}
  \And Yunfeng Shao\\
  Huawei Noah’s Ark Lab\\
  \texttt{shaoyunfeng@huawei.com}
   \And Jian Li \\
  Huawei Intelligent Automotive Solution BU\\
  \texttt{brandon.lijian@huawei.com}
}

\begin{document}
\newtheorem{definition}{Definition}
\maketitle

\begin{abstract}
Federated learning (FL) was proposed to achieve collaborative machine learning among various clients without uploading private data. However, due to model aggregation strategies, existing frameworks require strict model homogeneity, limiting the application in more complicated scenarios. Besides, the communication cost of FL's model and gradient transmission is extremely high. This paper proposes Loosely Coupled Federated Learning (LC-FL), a framework using generative models as transmission media to achieve low communication cost and heterogeneous federated learning. LC-FL can be applied on scenarios where clients possess different kinds of machine learning models. Experiments on real-world datasets covering different multiparty scenarios demonstrate the effectiveness of our proposal.
\end{abstract}

\section{Introduction}
\label{intro}

Increasing concern of user privacy drove laws like GDPR to prescribe proper usages of private data. Hence, federated learning (FL) emerged as a promising tool to implement machine learning tasks while keeping distributed data localized \cite{mcmahan2016communication}. Current FL methods have been employed on several applications with promising performance \cite{hard2018federated,yang2018applied}, revealing its potential in broader realistic scenarios.

However, several drawbacks among existing methods limit the further development of FL. Firstly, traditional FL structure based on model aggregation assumes homogeneity among all clients (i.e. every involved local model should be identical), which cannot always be met in applications \cite{sahu2018convergence}. Secondly, the exchange of model parameters results in extremely high communication cost \cite{li2019convergence}. Besides, conventional methods treat all clients the same\cite{yang2019federated}, calling for new methods able to measure client contributions. To this end, researchers tend to find alternative solutions for heterogeneous FL. But most of them are still based on parameter/gradient aggregation \cite{sen2019think,gao2019hhhfl}, which limits the degree of heterogeneity. Besides, the high cost of communication transmission is still hard to reduce, especially for edge devices \cite{wang2019adaptive}.

This paper proposes Loosely Coupled Federated Learning (LC-FL), a FL framework capable of solving highly heterogeneous model FL problems. For each client, our method can almost implement federated learning process regardless of machine learning methods. With the involvement of data generators, LC-FL generates artificial data that possesses both knowledge for model training and privacy guarantee for federated learning. With a data selection criterion and model training algorithm, generated data is used for local model updating with maximum efficiency.

Our contributions are twofold. Firstly, during learning process, proposed LC-FL enables highly heterogeneous models at each client instead of keeping model identity, which means the method can be applied in much more complicated scenarios. Secondly, LC-FL diverts model updating process from local clients to server side and reduces communication cost by avoiding iterative model/gradient transmission.

The remainder of paper is organized as follows. Section 2 introduces background and related work of related techniques. Section 3 describes LC-FL method, together with theoretical justification. Section 4 reports experiment results. Finally, we conclude the full paper in Section 6.  

\section{Background}
\subsection{Federated learning}
Federated learning (FL) algorithms aim to assign machine learning tasks to multiple remote clients with their own data. Unlike distributed learning, FL keeps data locally on each client instead of uploading or broadcasting it \cite{mcmahan2016communication}.

Based on data distribution variances, FL can be classified as horizontal, vertical and transfer FL \cite{yang2019federated}. This paper focus on horizontal FL, which means features of data among clients are identical while the specific data is different from each other.

As privacy limitation is strictly executed in FL, traditional methods tend to transmit alternative knowledge carriers such as model parameters and gradients. Google proposed FedAvg (Federate Averaging) in 2016 \cite{mcmahan2016communication}. In FedAvg, server broadcasts an initial model to clients. Clients update received models with local data and upload them back. With an model aggregation process, a global model is combined for later training loop. 

FedAvg is an efficient algorithm for FL settings especially when data is non-i.i.d. \cite{li2019convergence}. Nowadays, all commercial FL projects are using FedAvg-like structures. For example, Google and Apple are using FL in next-word prediction separately on GBoard and QuickType \cite{hard2018federated, kairouz2019advances}. There are also FL applications in financial and clinical scenarios \cite{federatedai_2020, brisimi2018federated}.However, FedAvg-like methods require client models to be exactly the same. Besides, parameters/gradients are transmitted in every global learning epoch, causing extremely high communication cost \cite{konevcny2016federated}. To overcome these shortages, a possible way is to introduce data generators in training, which carry data distribution knowledge and only need to be transmitted once during learning epochs.

\subsection{Generative models}
Data generator is a model generating artificial data that obeys original data distribution \cite{jaakkola1999exploiting}. As heterogeneous FL rejects aggregation-based updating because local models are not identical, this paper suggests transmitting generative models instead of model parameters/gradients. Hence, several data generator paradigms are introduced and applied in LC-FL.

The first introduced generator is VAE (Variational Auto Encoder) \cite{kingma2013auto}. Compare with traditional auto-encoder, by restricting hidden variables' distribution, VAE can generate artificial data once fed with Gaussian noise \cite{pu2016variational}.Another generator this paper referred to is GAN (generative adversarial network) \cite{goodfellow2014generative}. GAN can effectively generate synthetic data with high-accuracy details through adversarial training. However, coming with the details is GAN's high computation and generation cost \cite{genevay2017gan}, making it quite time-consuming.

Another perspective of data generation is data synthesis. Here we introduce dataset distillation, a method aiming to compress a huge dataset into a limited one while maintaining training performance on neural networks \cite{wang2018dataset}. Though dataset distillation is not a generation approach, it can also produce knowledge-carrying and privacy-free data. 

In summary, data generator types can be various. Based on different requirements and data types, a proper generator will be chosen to solve specific tasks. 

\section{Methods}
In this section, LC-FL, a heterogeneity-adapted FL method based on artificial data, is introduced. Before introducing the implementation of LC-FL, we first formally define the loosely coupled FL scenario in this paper. 

\subsection{Problem setup}
Federated learning typically aims to minimize the loss function of client models on global test sets. However, considering the heterogeneity of client models, this paper defines a highly heterogeneous model FL problem called loosely coupled FL. For clients, model structures and even model approaches can be variant. For example, some clients may own different neural networks as classifiers while others obtain SVM, GBDT or Logistic Regression models. In detail, only two characteristics are required to achieve FL training: firstly, all models involved should receive inputs and outputs of the same dimension. Besides, they should provide probabilities or confidence levels for classification. These two requirements are easy to meet, bringing high adaptability to our method.

\begin{figure}
	
	\centering
	\includegraphics[width=1\textwidth]{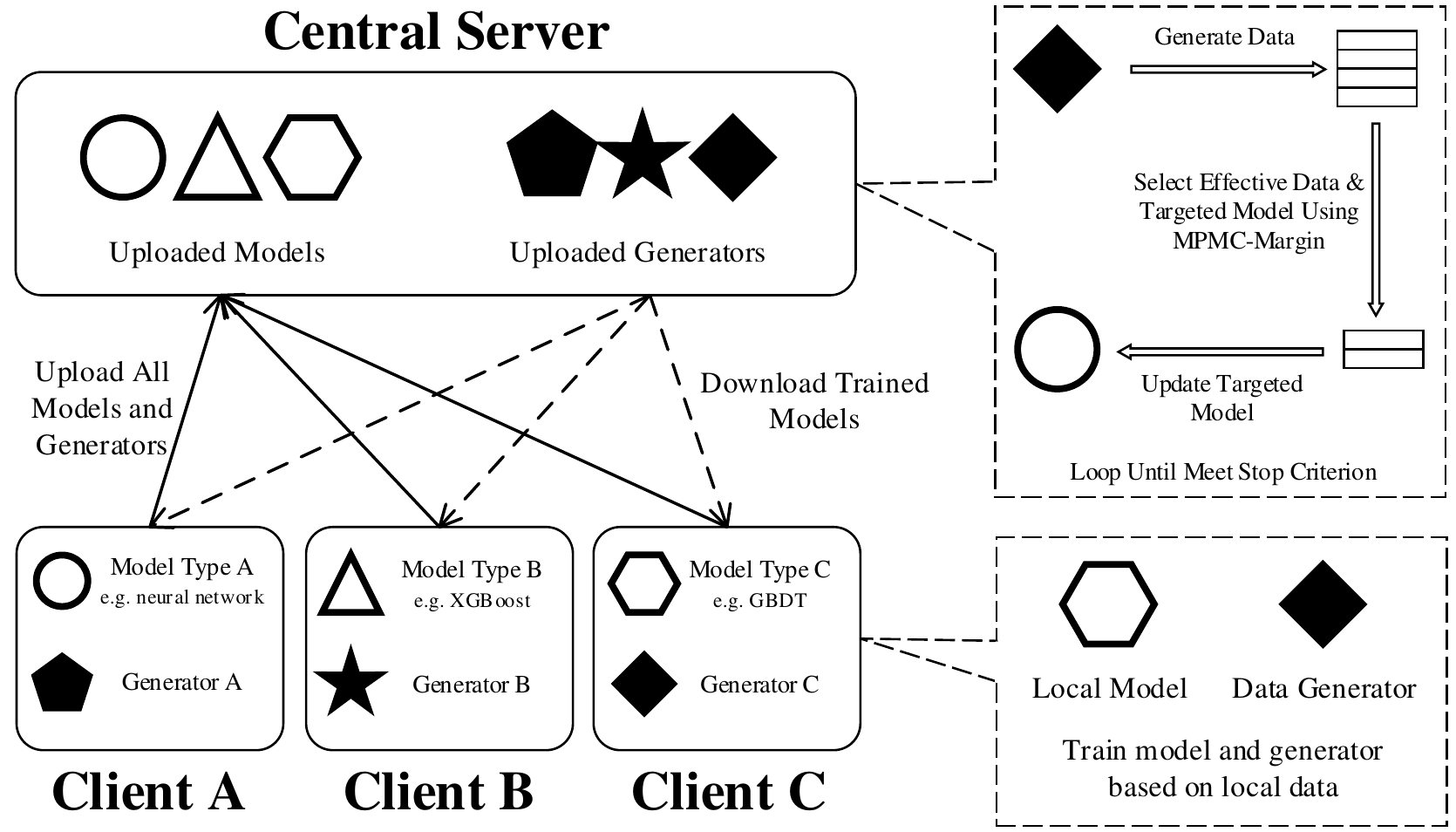}
	\caption{Loosely Coupled Scenario and Mainframe of LC-FL}
	\label{fig1}
\end{figure}

The scenario of loosely coupled FL is a typical server-client configuration. A central server connects all other clients and is responsible for most model updating issues. Among clients, a loosely coupled relationship is applied. Clients can't acquire each other's model and data status only if they met the requirements mentioned above.  Notably, LC-FL can support decentralized scenarios after minor adjustment. But this paper just considers server-client situations.

\subsection{LC-FL}
\subsubsection{Mainframe}
This paper proposed a method called LC-FL to solve this scenario. In this method, we separate the learning progress into two parts. One is at the client side and the other is at server side. The concrete process is shown in Fig \ref{fig1}.

\begin{algorithm}[h]
	\caption{LC-FL}
	\begin{algorithmic}[1]
		\REQUIRE 
		Clients $1,2,...,n$, each owns a local private dataset $S_{i}$, a local model $h_{i}$ and a data generator $G_{i}$. \\
		A center server $C$. Training iteration limitation $N$.
		\ENSURE 
		Trained local models $h_{1},...,h_{n}$.\\
		\hspace*{-0.2in}{\bf Procedure:} 
		\STATE All clients train and upload their local model and data generator to center server.
		\STATE Center server uses each generator to generated an artificial dataset $D_{i}$.
		\STATE Inner iteration counter $T=0$.
		\WHILE{$T<N$}
		\STATE Sample a client $i$ according $|S_{i}|/\Sigma^{n}_{i=1}|S_{i}|$.
		\STATE Client $i$ randomly selects an example $(x,y)\in D_{i}$.
		\STATE Client $i$ computes MPMC-margin $\rho_{H}(x,y)$ according to (\ref{eq1}).
		\STATE Records the client $i^{+}$, $i^{-}$ and maximum incorrect class $y^{-}$ as in (\ref{eq2}).
		\IF{$\rho_{H}(x,y)\leq0$}		
		\IF{$i^{+}\neq i~or~i^{-}\neq i$}
		\STATE Center server $C$ sends $(x,y,y^{-})$ to $i^{+}$ and $i^{-}$.
		\STATE Client $i^{+}$ and $i^{-}$ add received data to training set and update model.
		\STATE $T=T+1$.
		\ENDIF
		\ENDIF
		\ENDWHILE		
	\end{algorithmic}
	\label{alg1}
\end{algorithm}

The pseudo algorithm code of LC-FL is shown in Alg \ref{alg1}. The first part is pre-training in all participants. In this part, every client trains a local model and a data generator on its own data. With regard to data generator, after meeting performance demands, privacy protection processing is also required to avoid malicious inspection from server-side. For example, confusion and encryption methods can be effective ways to enhance privacy level. In application, pre-training operations should be done before FL process.

The second part is server-side model updating. After receiving local models and data generators from clients, firstly, server checks and uses generators to produce an amount of artificial data. Unlabeled artificial data is also classified with local models. Based on an assigned selection criterion (this paper selects with the probability based on client local data amounts), server chooses a piece of synthetic data. Through a updating criterion called MPMC-Margin, server decides whether this data is beneficial to updating some participants' models. Then server collects valuable data to update specific models. This selection-updating process is executed in loops until reaching stopping conditions. 

After server-side updating, the final part is local fine-tuning for all participants. Clients download their own updated models from central server. Based on local private data, client fine tunes model with prudent strategy (e.g. elastic weight consolidation). As a result, clients obtain models both suitable for local and global data classification.

\subsubsection{MPMC-Margin}
In server-side model updating process, an essential part to boost updating speed is filtering efficient data. Hence this paper introduces MPMC-Margin (multi-party multi-class margin) as data selection criterion.

MPMC-Margin is inspired from heterogeneous model reuse (HMR), a distributed learning algorithm \cite{wu2019heterogeneous}. It's defined on a set of local models $H=\{h_{1},...,h_{n}\}$ at a labeled data $(x,y)$ as:
\begin{equation} \label{eq1}
\begin{split}
\rho_{H}(x,y)&=\max_{i}h_{i}(x,y)-\max_{j,y'}h_{j}(x,y'),\\
where~y&\in \mathcal{Y}_{i},y' \in \mathcal{Y}_{i}\setminus\{ y \}.
\end{split}
\end{equation}
We interpret MPMC-Margin with a plain understanding. If a local model classifies a sample by mistake with a high confidence, we can legitimately assume that this model may not obtain proper separating hyperplane for such samples. Model performance will benefit a lot by involving such data into training set. Besides, for local models who correctly classify samples with highest confidence, this sample can also be applied to enhance the hyperplane as well.

As a result, for every chosen artificial data, its MPMC-Margin values are computed on all local models. Then LC-FL selects two participants $i^{+},i^{-}$ via selection criterion:
\begin{equation} \label{eq2}
\begin{split}
i^{+}&=\mathop{\arg\min}_{i}h_{i}(x,y),where~y \in \mathcal{Y}_{i},\\
(i^{-},y^{-})&=\mathop{\arg\min}_{j,y'}h_{j}(x,y'),where~y'\in \mathcal{Y}_{j}\setminus\{y\}
\end{split}
\end{equation}

Therefore, the MPMC-margin can also be represented as:  
\begin{equation} \label{eq2}
\begin{split}
\rho_{H}(x,y)=h_{i^{+}}(x,y)-h_{i^{-}}(x,y^{-})
\end{split}
\end{equation}   

Further, this paper defines MPMC-margin loss to evaluate the overall margin status of all generated data. The loss is defined as:

\begin{equation} \label{eq2}
\begin{split}
R_{S}(H)&=\frac{1}{|S|}\sum_{(x,y)\in S}\ell(\rho_{H}(x,y))\\
\ell(p)&=\left\{
\begin{array}{rcl}
1  &if~\rho \leq 0,\\
0  &if~\rho > 0.
\end{array} 
\right.
\end{split}
\end{equation}
LC-FL's goal of model updating is converted to minimize MPMC-margin. This can be achieved by increasing small margins. Both increasing $h_{i^{+}}$ or decreasing $h_{i^{+}}$ will help.

\subsubsection{Model updating strategy}
After data selection, next step is to update local models. To reduce communication cost, all updating operations are transferred from client to server side with the help of data generator. For a specific client, when selected data amount reaches a threshold set both considering training impact and selection time, these samples will be mixed with artificial data this client already owns. Then server updates this model for several epochs to ensure new hyperplane is built. 

Notice that all selected artificial data will remain in this client's dataset till updating phase finishes. This will ensure that as the updating process carrying on, more generated data will be added to improve the global performance of local models. 

\subsubsection{Contribution evaluation}
This paper also provides an algorithm to evaluate the contribution of each client quantitatively. Specifically, due to our data selection and model updating strategy, each selected piece of data has generator ID and transmitted client ID recorded. These records can be applied to measure quality of both local models and generators. Such contribution evaluation can help encourage clients carrying effective generators to join FL process for profit. Its detailed information is introduced in Supplementary Files. 

\subsection{Privacy concern}
An extensively concerned factor in data generator is privacy. In FL, two types of attacks should be considered: transmission-phase spying and server-side inspection \cite{sallette2000system}. LC-FL achieves privacy-preserved FL based one assumption: server is roughly credible as it undertakes most calculation processes. In the worst case, it's curious but not malicious. Based on this assumption, two detailed measures are implement. During transmission phase, both symmetric and asymmetric encryption algorithms can be applied. As for spy-disposed server-side inspectors, LC-FL can use confusion-based approaches. Differential privacy is an ideal technique. 

Here we take GAN as a generative model example. The conception of differential privacy has been introduced and applied in many FL-related generative methods \cite{jordon2018pate}. In LC-FL, user-level DP proposed by DP-FedAvg \cite{augenstein2019generative} and FedGP \cite{triastcyn2019federated} can be applied. Here the DP can be formally defined as:

\begin{definition}
Differential Privacy: A randomized mechanism $\mathcal{M}:\mathcal{D}\rightarrow\mathcal{R}$with domain $\mathcal{D}$ and range $\mathcal{R}$ satisfies $(\epsilon, \delta)$-differential privacy if for any two adjacent inputs d, $d'\in \mathcal{D}$ and for any subset of outputs $\mathcal{S} \subseteq \mathcal{R}$ it holds that:
$$
Pr[\mathcal{M}(d)\in\mathcal{S} \leq e^{\epsilon}Pr[\mathcal{M}(d')\in\mathcal{S}]+\delta
$$
\end{definition}
Then Gaussian noise mechanism defined as follows is chosen:
$$
\mathcal{M}(d) \triangleq f(d)+\mathcal{N}(0,s^{2}_{f}\cdot\sigma^{2})
$$
This can help improve privacy bounds. Adding such a noise layer in the discriminator of GAN, a same DP level can be guaranteed in generator side \cite{triastcyn2019federated}. As for VAE and dataset distillation, similar DP method can be applied to ensure privacy level during FL \cite{chen2018differentially,wang2018dataset}. Based on the application circumstances of LC-FL, this paper also assumes server has the power to avoid external hacks, which means user-level differential privacy is enough for basic privacy guarantee.   

\section{Experiments}
This paper will evaluate LC-FL on different scenarios. Three tasks are mainly applied to demonstrate its performance. First experiment demonstrates that data generator can replace real data considering training performance. Then LC-FL is compared with benchmark FL approaches. At last, a heterogeneous-model experiment is deployed to show the generalization ability of LC-FL. 

\subsection{Experiment setting}
To simulate various FL settings, Two types of scenarios (shown in Table \ref{table1}) are set to represent different data distribution situations. Detailed information of all involved scenarios can be found in Supplementary Files.

\begin{table}
	\caption{Federated learning scenarios for LC-FL}
	\centering
	\begin{tabular}{ccccccc}
		\toprule
		\multicolumn{3}{c}{Homogeneous(7 clients)} & \multicolumn{3}{c}{Heterogeneous(8 users)}  \\
		\cmidrule(r){1-3} \cmidrule{4-6}
		Scenario     & Classes per client & Data per class    & Scenario     & Classes per client & Data per class  \\
		\midrule
		A & 10 & Equal & A & 10 & Equal\\
		B & 10 & Unequal & B & 3-4 & Unequal\\
		C & 8-9 & Unequal &  &  & \\       
		D & 2-4 & Unequal &  &  & \\ 
		\bottomrule
	\end{tabular}
	\label{table1}	
\end{table}

First type is homogeneous scenario in which all clients share the same neural network structures. As data distribution varies, LC-FL's performance is examined under both i.i.d. and non-i.i.d. situations. Another type is heterogeneous scenario where clients posses different types of models. This paper uses two neural network structures(CifarNet and a modified version of MNISTNet) and two non-NN machine learning methods(logistic regression and XGBoost). Under such scenarios, LC-FL will demonstrate its superior capability of supporting highly heterogeneous FL tasks. Notably, in all scenarios, each training sample is obtained by only one client.

Three kinds of data generators are tested in experiments, among which dataset distillation and VAE are only applied in generated data comparison section while GAN is used in all examination sections. All data generators mentioned are trained on each client with local data. 

This paper involves two common public datasets. FASHION-MNIST, a MNIST-like dataset containing clothing images is applied to compare the performance of real and artificial data \cite{xiao2017fashion}. CIFAR-10 is chosen as test set to acquire more comparable results in other experiments \cite{krizhevsky2009learning}. 

\subsection{Comparison between real and artificial data}
To prove that artificial data can act in a same manner as real data during FL training, this experiment applies different types of artificial data in LC-FL respectively for comparison with real data performance. As for metrics, the performance of different data can be evaluated by their ability to improve a same local model's test accuracy. The results are shown in Table \ref{table2}.

\begin{table}
	\caption{LC-FL test accuracy on different data type and scenarios (Fashion-MNIST)}
	\centering
	\begin{tabular}{cccccc}
		\toprule
		 Scenario   & Initial Accuracy &Original data & Distilled data    & VAE     & GAN   \\
		\midrule
		
		A & 0.8816 & 0.9004 & 0.9024 & \textbf{0.9043} & 0.9021\\
		B & 0.8795 & 0.8993 & 0.9003 & \textbf{0.9028} & \textbf{0.9028}  \\       
		C & 0.8481 & 0.8907 & 0.8475 & 0.8731 &  \textbf{0.8792} \\ 
		D & 0.3510 & 0.8209 & 0.6445 & 0.7218 & \textbf{0.8164}\\
		\bottomrule
	\end{tabular}
	\label{table2}	
\end{table}

In conclusion, all generators can significantly improve local models' average classification accuracy on global test set. Among all tested approaches, GAN outperforms others especially on extremely unbalanced scenarios. Besides, compared with true data, GAN-generated data obtains almost the same performance. This indicates that generated data can effectively replace original one in federated learning. In later experiments, only GAN is used as generative model paradigm. 

\subsection{Algorithm performances in homogeneous scenarios}
Having validated the practicability of artificial data, next goal of LC-FL is to check its performance compared with conventional federated learning methods like FedAvg and FedProx.

Performances are demonstrated in three metrics: global test accuracy, training time and model transmission cost. Among these metrics, model transmission cost means the number of models transferred to achieve full workflow. Here we assume generators have same size as local models for simplification. To ensure fairness, a same neural network, CifarNet, is applied among all involved approaches. As LC-FL updates two clients each training step, we also set updating ratio of FedAvg and FedProx to 0.3 to make comparison equivalent. The details of model structures is introduced in supplementary files. And after a 10-times experiment, results are shown in Fig \ref{fig2} and Table \ref{table3} below. 
\begin{figure}
	\centering
	\subfigure[Scenario A]{
		\includegraphics[width=0.48\textwidth]{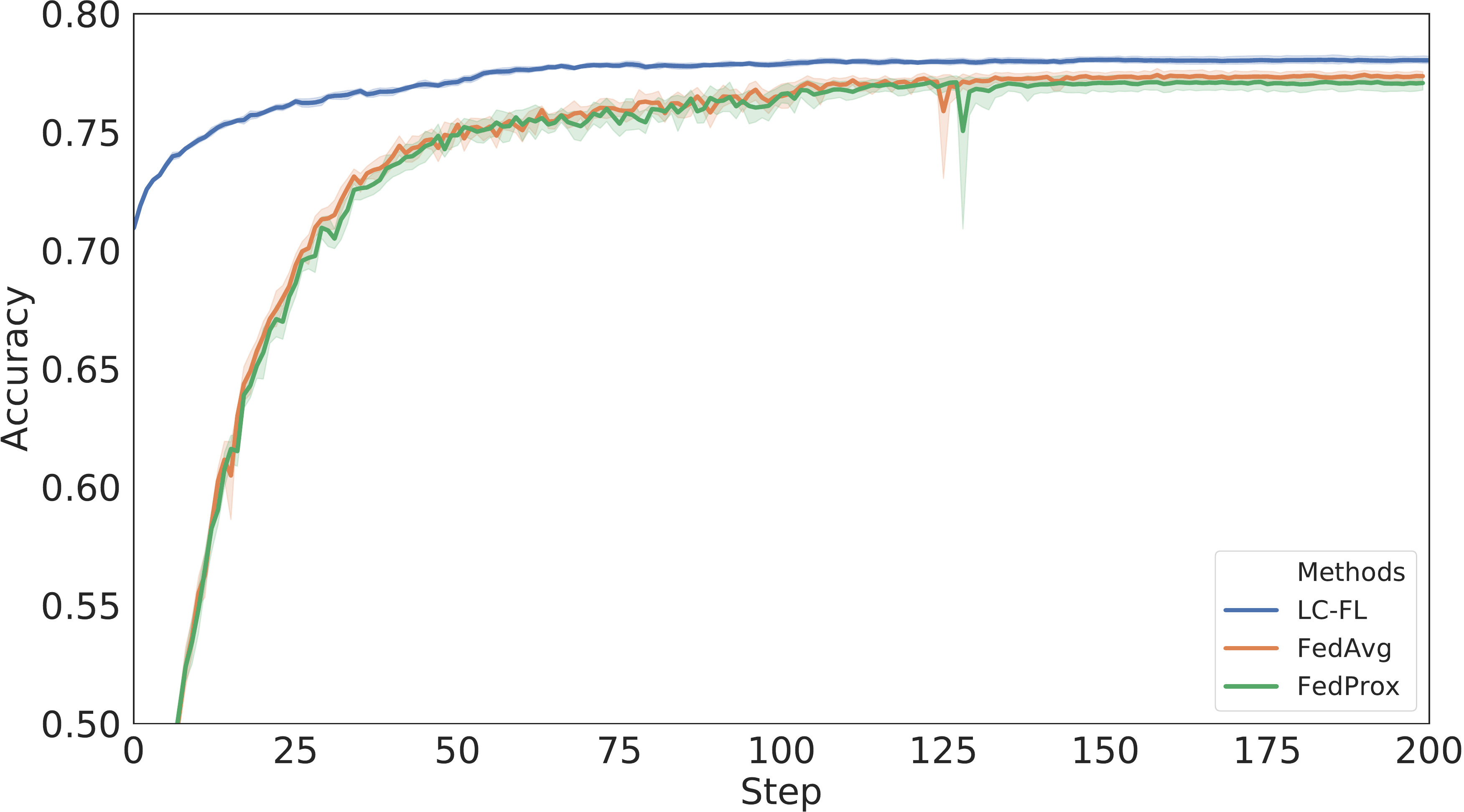}	
		\label{fig2a}	
	}
	\subfigure[Scenario B]{
		\includegraphics[width=0.48\textwidth]{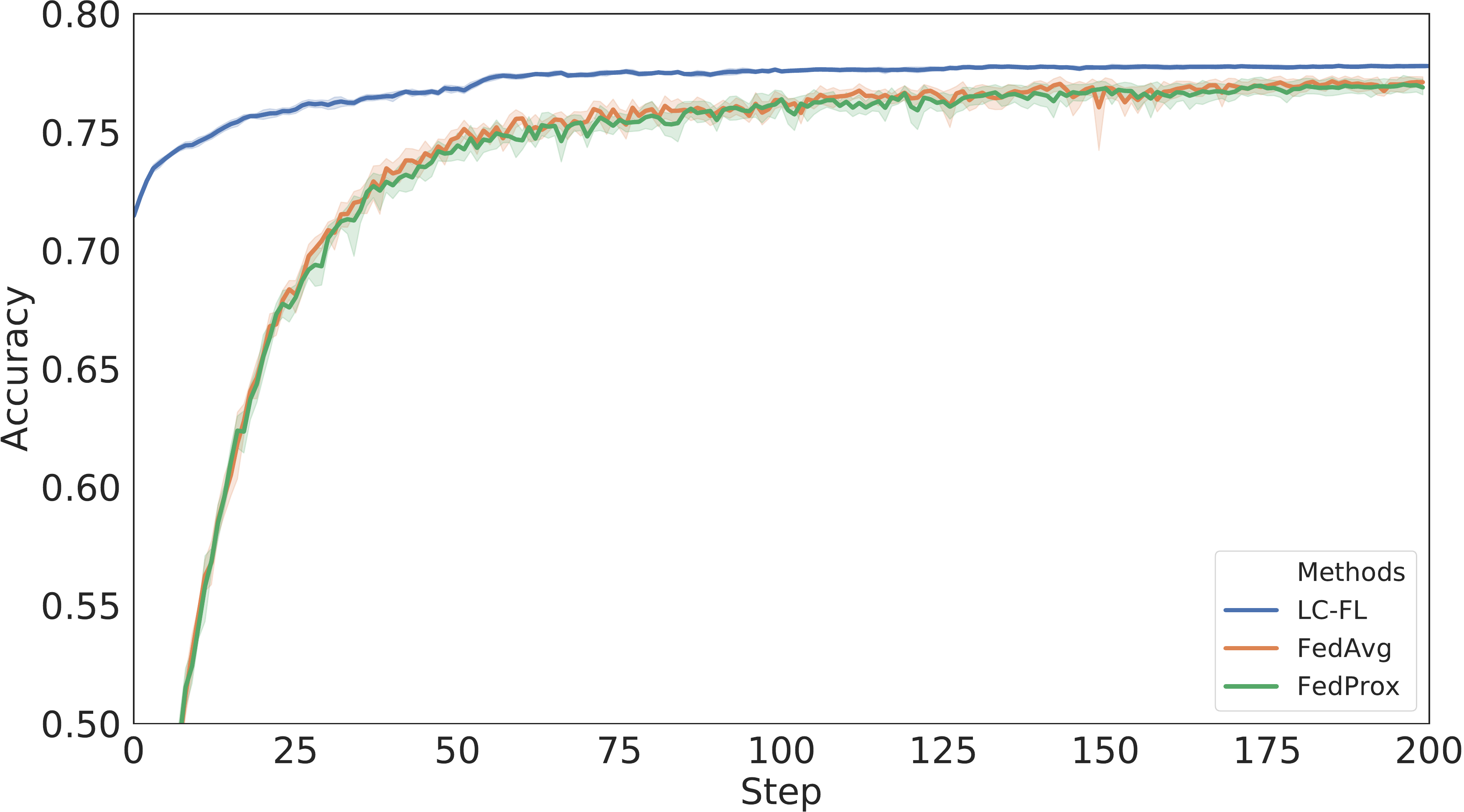}
		\label{fig2b}	
	}
	\subfigure[Scenario C]{
		\includegraphics[width=0.48\textwidth]{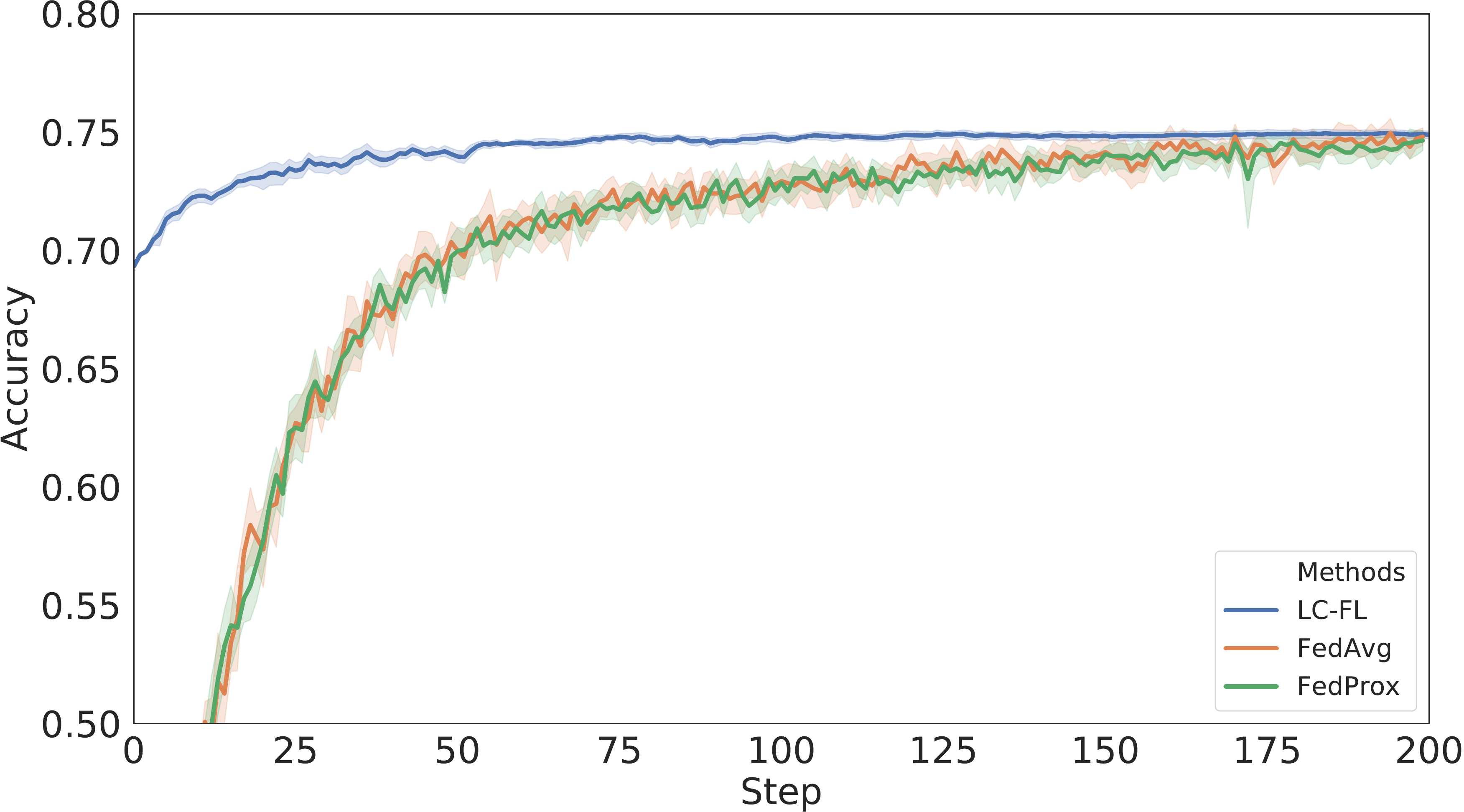}
		\label{fig2c}	
	}
	\subfigure[Scenario D]{
		\includegraphics[width=0.48\textwidth]{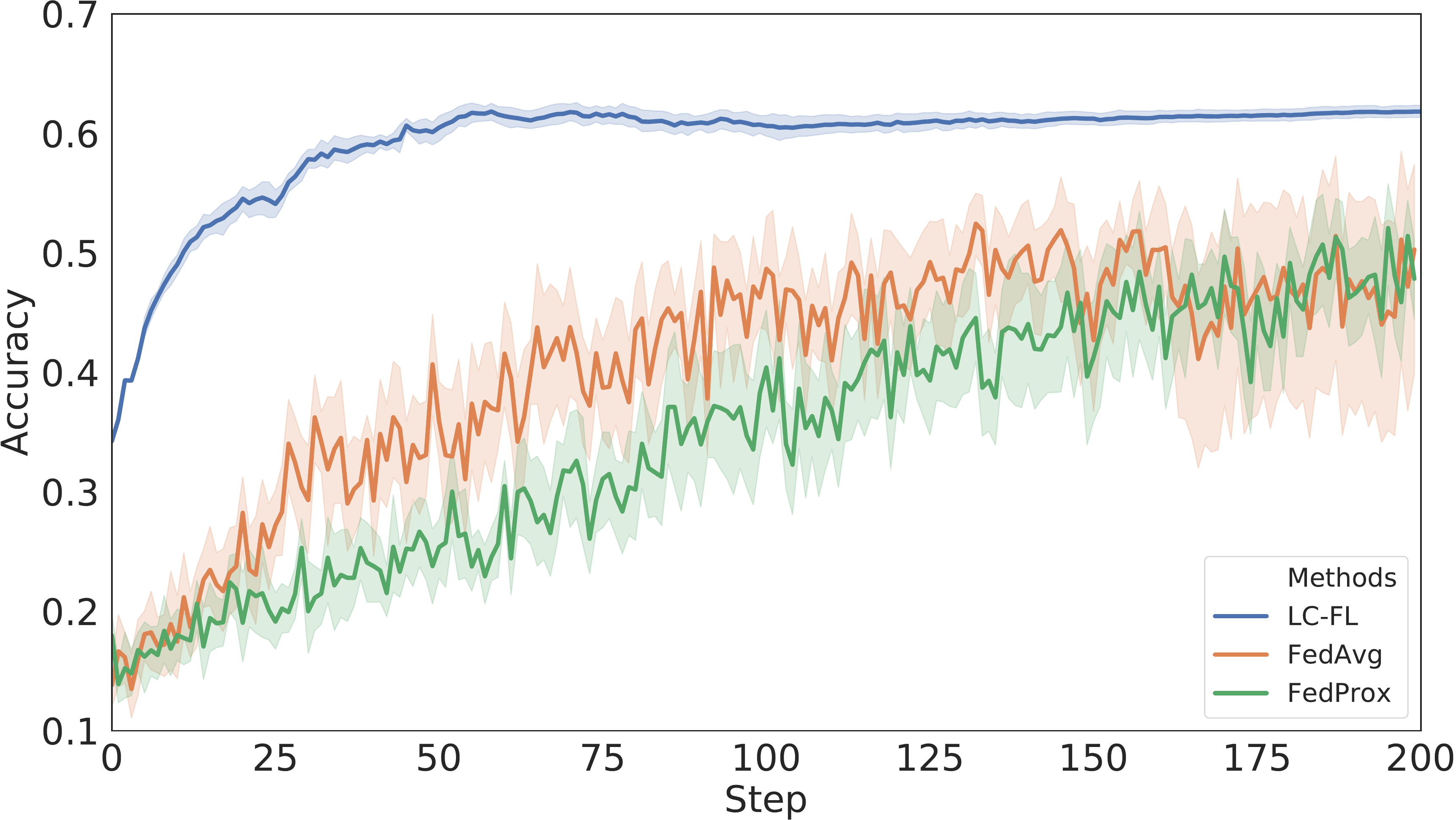}
		\label{fig2d}	
	}
	\caption{Homogeneous Federated Learning Performances (CIFAR-10).}
	\label{fig2}	
\end{figure}

\begin{table}
	\caption{FL methods under homogeneous scenarios}
	\label{sample-table}
	\centering
	\renewcommand{\arraystretch}{1.2}
	\begin{tabular}{ccccc}
		\hline
		Scenario & Method & Test Accuracy & Training Time (min) & Model Transmission \\
		\hline		
		\multirow{3}*{A} & LC-FL & \textbf{0.7804 $\pm$ 0.0025} & $229\pm16.1$ & \textbf{21}\\
		& FedAvg & $0.7736\pm0.0031$ & \textbf{96.4 $\pm$ 2.3} & 1800\\
		& FedProx & $0.7707\pm0.0049$ & $163.4\pm2.5$ & 1800\\
		\hline
		\multirow{3}*{B} & LC-FL & \textbf{0.7779 $\pm$ 0.0012} & $252.7\pm22.4$ & \textbf{21}\\
		& FedAvg & $0.7711\pm0.0034$ & \textbf{96.0 $\pm$ 2.0} & 1800\\
		& FedProx & $0.7689\pm0.0054$ & $163.1\pm1.3$ & 1800\\
		\hline
		\multirow{3}*{C} & LC-FL & \textbf{0.7491 $\pm$ 0.0027} & $237.2\pm26.8$ & \textbf{21}\\
		& FedAvg & $0.7481\pm0.0027$ & \textbf{97.9 $\pm$ 2.4} & 1800\\
		& FedProx & $0.7465\pm0.0070$ & $165.0\pm1.9$ & 1800\\
		\hline
		\multirow{3}*{D} & LC-FL &\textbf{0.6182 $\pm$ 0.0083}  & $169.5\pm16.1$ & \textbf{21} \\
		& FedAvg & $0.5028\pm0.1515$ & \textbf{96.5 $\pm$ 2.8} & 1800\\
		& FedProx & $0.4783\pm0.0537$ & $163.7\pm1.7$ & 1800\\
		\hline
	\end{tabular}
	\label{table3}
\end{table}

In terms of accuracy, LC-FL outperforms other two methods in all scenarios along with much higher convergence speeds. Besides, LC-FL reduces communication cost by an order of magnitude, making it quite practical when transmission capability is severely restricted (e.g. complex IoT scenes). Notably, in contrast with parameter-server based FL approaches, LC-FL needs more time to complete same epochs of training. There are two possible explanations. First, LC-FL's training time includes generating serial local initial models and generative models, which takes quite a while. Second, our experiments use array migration to simulate transmission process, making transmission time almost zero for all tested methods, which gives benefits to conventional FL approaches.

In summary, with generative models and efficient data selection criterion, LC-FL possesses great capability achieving effective FL on both i.i.d. and non-i.i.d. scenarios.

\subsection{Algorithm performances in heterogeneous scenarios}
\begin{figure}
	\centering
	\subfigure[Scenario A]{
		\includegraphics[width=0.48\textwidth]{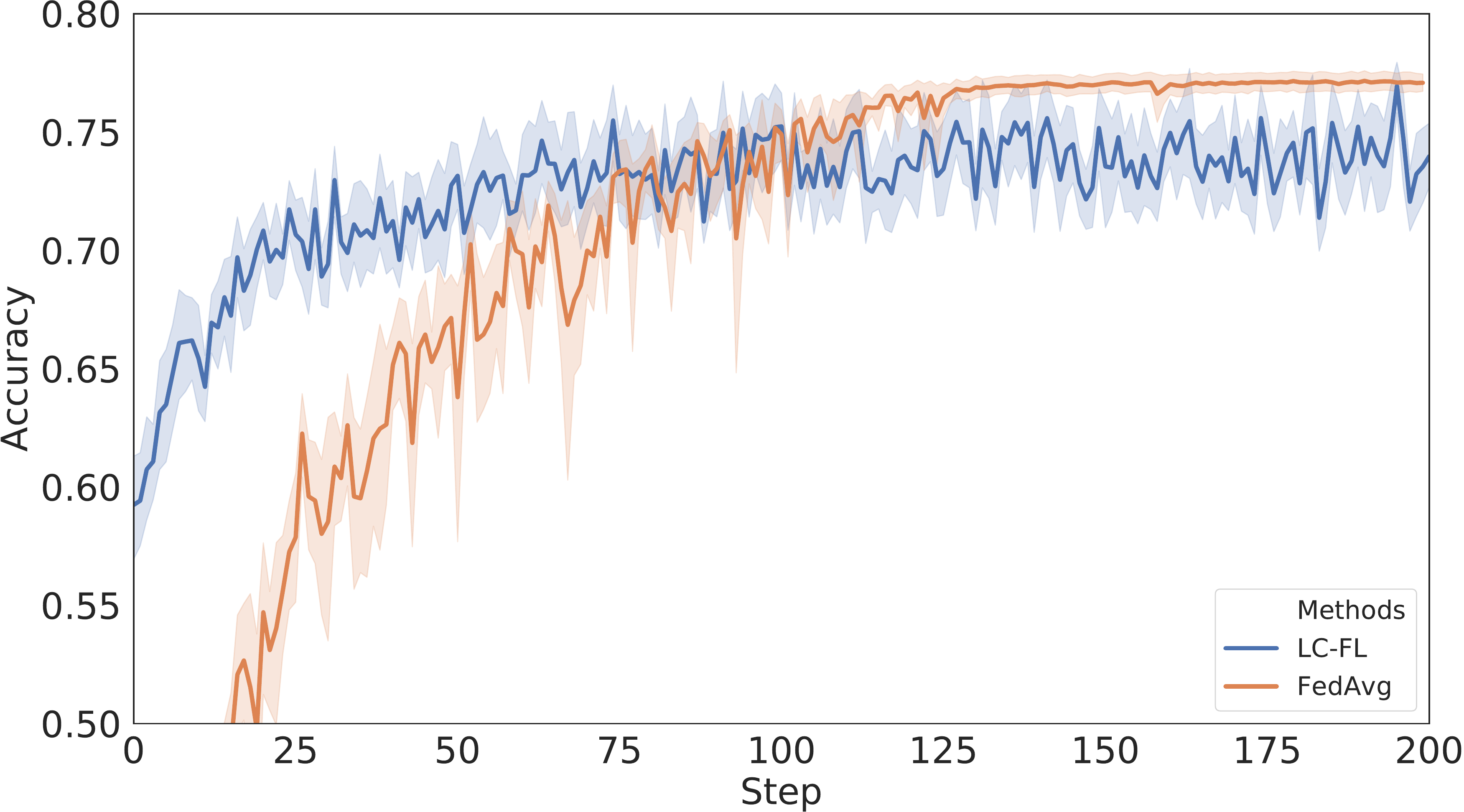}
		\label{fig31}	
	}
	\subfigure[Scenario B]{
		\includegraphics[width=0.48\textwidth]{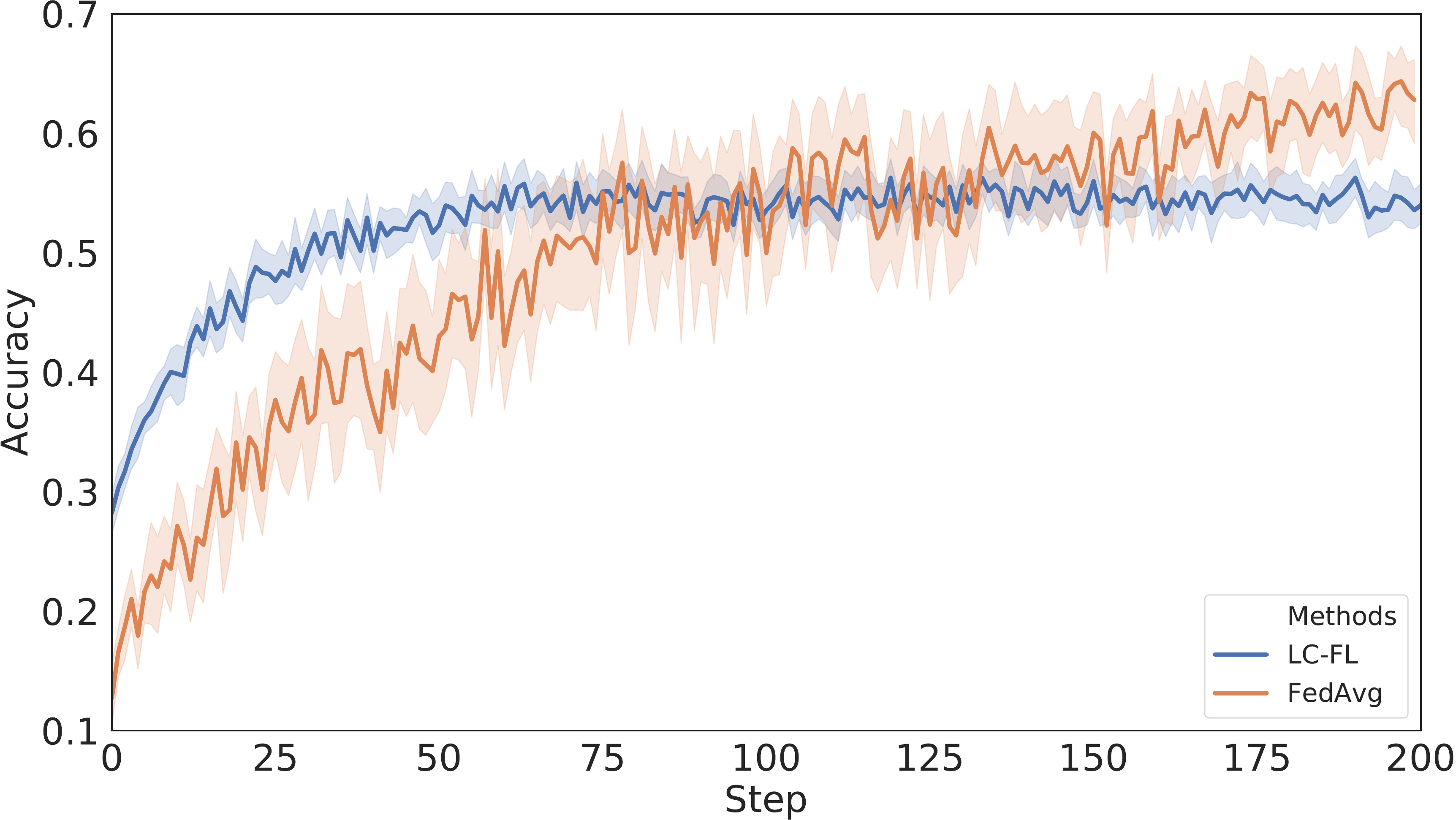}
		\label{fig32}	
	}
	\caption{Heterogeneous federated learning performances (CIFAR-10).}
	\label{fig3}	
\end{figure}

\begin{table}
	\caption{FL methods under heterogeneous scenarios}
	\label{het-table}
	\centering
	\renewcommand{\arraystretch}{1.2}
	\begin{tabular}{cccc}
		\hline
		Scenario & Method & Test Accuracy & Model Transmission \\
		\hline
		\multirow{2}*{A} & LC-FL & $0.7354\pm0.0236$ & 24\\
		& FedAvg & $0.7708\pm0.0058$ & 2000\\
		\hline
		\multirow{2}*{B} & LC-FL & $0.5358\pm0.0245$ & 24 \\
		& FedAvg & $0.6270\pm0.0576$ & 2000\\
		\hline
	\end{tabular}
	\label{table4}	
\end{table}
Apart from implementing FL effectively, another property of LC-FL is its supporting heterogeneous machine learning methods. Hence LC-FL is also tested on two such situations. Though some clients owns non neural network methods, we still implement FedAvg for comparison. Besides, As two traditional machine learning methods need long time to train, only two metrics are demonstrated: test accuracy and model transmission cost. We also consider all local models and generators have the same size for simplification. Results are shown in Fig \ref{fig3} and Table \ref{table4}. 

Similar to results above, based on generative models and related synthetic data, LC-FL can effectively improve global test accuracy for all clients. This experiment demonstrate the generalization ability of LC-FL, which might bring broader application space for federated learning.

\section{Conclusion}
This paper described loosely coupled federated learning scenario where conventional FL methods are invalid. Hence, we proposed LC-FL, a workflow over generative models for such heterogeneous scenarios. Using artificial data and an MPMC-Margin based selection criterion, LC-FL got excellent performances in global test accuracy and transmission cost, along with fantastic generalization ability supporting heterogeneous machine learning methods in a same FL task.

Except mentioned generative models, RNN-based networks are also commonly-used for data generation. In future, we consider explore more sequence related FL tasks, adding applicability to LC-FL.

\end{document}